\documentclass[letterpaper, 10pt, conference]{ieeeconf}

\IEEEoverridecommandlockouts
\usepackage{amsmath,amsfonts,amssymb}
\usepackage{graphicx}
\usepackage{subcaption}
\usepackage[colorlinks=true, allcolors=black, urlcolor=magenta]{hyperref}
\usepackage{algorithm}
\usepackage[noend]{algpseudocode}
\usepackage[noadjust]{cite}
\usepackage{multirow}
\usepackage{colortbl}
\usepackage[dvipsnames]{xcolor}
\usepackage{fancyhdr}

\fancypagestyle{firstpage}{%
 \fancyhf{}%
 \fancyheadoffset{0cm}%
 \fancyhead[L]{\vskip -1cm 2024 International Conference on Robotics and Automation (ICRA 2024)\newline
  \textit{May 13-17, 2024. Yokohama, Japan}}
}

\title{Gradient-based Local Next-best-view Planning for Improved Perception of Targeted Plant Nodes}
\author{Akshay K. Burusa$^{1}$, Eldert J. van Henten$^{1}$, Gert Kootstra$^{1}$%
\thanks{This research was funded by the Netherlands Organization for Scientific Research (NWO) project FlexCRAFT, grant P17-01}%
\thanks{$^{1}$ Agricultural Biosystems Engineering, Wageningen University and Research, The Netherlands {\tt\small akshaykumar.burusa@wur.nl}}%
}

\begin{document}

\maketitle
\thispagestyle{firstpage}

\begin{abstract}
Robots are increasingly used in tomato greenhouses to automate labour-intensive tasks such as selective harvesting and de-leafing. To perform these tasks, robots must be able to accurately and efficiently perceive the plant nodes that need to be cut, despite the high levels of occlusion from other plant parts. We formulate this problem as a local next-best-view (NBV) planning task where the robot has to plan an efficient set of camera viewpoints to overcome occlusion and improve the quality of perception. Our formulation focuses on quickly improving the perception accuracy of a single target node to maximise its chances of being cut. Previous methods of NBV planning mostly focused on global view planning and used random sampling of candidate viewpoints for exploration, which could suffer from high computational costs, ineffective view selection due to poor candidates, or non-smooth trajectories due to inefficient sampling. We propose a gradient-based NBV planner using differentiable ray sampling, which directly estimates the local gradient direction for viewpoint planning to overcome occlusion and improve perception. Through simulation experiments, we showed that our planner can handle occlusions and improve the 3D reconstruction and position estimation of nodes equally well as a sampling-based NBV planner, while taking ten times less computation and generating $28\%$ more efficient trajectories.
\end{abstract}


\section{Introduction}

Robots are increasingly used in agro-food environments, such as tomato greenhouses, to meet the growing demand for food and to compensate for the growing labour shortage \cite{r2018research, oliveira2021advances}. Using robots, we can help automate labour-intensive tasks like harvesting and de-leafing in a greenhouse. To perform these tasks, robots must accurately and efficiently perceive the plant nodes, i.e., the plant parts that connect the leaves and fruits to the main stem. Detecting these nodes and localising their 3D position is essential to perform cutting and grasping action for harvesting and de-leafing. However, this is extremely challenging due to the high levels of occlusion in tomato plants, as the nodes are often hidden from the robot's view by leaves or other plant parts \cite{bac2017performance, kootstra2021selective}.

Active vision is a promising approach to handling occlusion, in which the robot reasons based on the information gathered so far and strategically moves the camera to better viewpoints to overcome occlusion and improve perception accuracy \cite{chen2011active}. Most active-vision algorithms are designed for global view planning, i.e. they aim to perceive an entire scene or a large structure. However, there are several tasks in a greenhouse that would benefit from a more local view planning that aims to perceive a specific plant part in more detail, such as inspection of fruits or pose estimation of cutting points. In such tasks, the goal is to plan viewpoints around a single target object and efficiently explore its properties. The widely-used approaches to global planning randomly sample a set of candidates to explore potential viewpoints for improving perception. Such global sampling-based approaches have some drawbacks for local planning: (i) a large number of candidates need to be sampled to explore the viewing space sufficiently, (ii) if insufficient candidates are sampled, it is highly likely that the actual best viewpoint is never sampled and hence never visited, and (iii) the trajectory of viewpoints generated may not be smooth and could have abrupt transitions across the viewing space. These drawbacks are more severe when the target object is highly occluded or the candidates are sampled with a greater degree-of-freedom (DoF) as the search space grows exponentially. As a result, sampling-based planning could result in longer processing times, higher computation costs, and an inefficient viewpoint trajectory for local planning.

In this paper, we aimed to address these drawbacks of sampling-based approaches for local viewpoint planning. We developed a gradient-based optimisation algorithm by using a differentiable utility function, which allowed us to move the camera such that the viewpoint utility was locally maximised. For the perception of a single target object, a local optima is desirable as it locally improves coverage and handles occlusion. We applied our method to the problem of 3D reconstruction and position estimation of plant nodes. Using a simulated environment with 3D mesh models of tomato plants, we evaluated the performance of our method and compared it against a sampling-based and a random method for local view planning. The main contributions of this paper are: (i) a novel gradient-based approach for local view planning to handle occlusions and improve perception of a target object, (ii) significant improvements in computational speed and efficiency of the generated viewpoint trajectories compared to sampling-based methods, and (iii) the evaluation and analysis of our method in simulation using tomato plant models with high levels of occlusion. Our code is available at \url{https://github.com/akshaykburusa/gradientnbv}.


\section{Related work}

Active vision is a well-researched topic \cite{placed2023survey}, especially for exploration or reconstruction of unknown scenes. Next-best-view (NBV) planning \cite{zeng2020view} is a popular approach to active vision, which plans one-step ahead. NBV planning methods can be categorised based on their exploration strategy.

Frontier-based methods identify the boundaries between known and unknown regions of the scene and plan viewpoints to explore these frontiers \cite{yamauchi1997frontier}. They are primarily formulated for exploration of large scenes and cannot be directly used for other perception tasks. Sampling-based methods generate a set of candidate viewpoints and select the one that maximises a utility function. Rapidly-exploring Random Tree (RRT) is popular method for sampling candidates \cite{bircher2016receding}. These methods allow for fast exploration of the scene and can be adapted to tasks such as localisation \cite{papachristos2017uncertainty} or object search \cite{kay2021semantically}. Hybrid approaches have also been proposed, that combine frontier and sampling-based methods to globally find frontiers and locally sample candidates around them \cite{meng2017two, dai2020fast}. These methods are designed for global planning and are not suitable for local or targeted NBV planning.

Gradient-based methods are an alternative approach, which use the gradient of some viewpoint utility function to guide the selection of the next-best viewpoint. They can be beneficial for local planning as they avoid the computational cost of sampling candidates and can generate a smooth sequence of viewpoints. In \cite{lehnert20193d}, it was shown that computing the gradient based on the surface area of a target object and servoing the camera along the gradient improved the target perception. However, they used a special multi-camera setup to compute the gradients, making their method difficult to apply on other platforms. Also, they relied only on the current sensor data and did not explicitly map the scene. In \cite{deng2020frontier}, a fuzzy logic filter was used to make a frontier-based utility function differentiable. This allowed the refinement of a global plan using gradient-based optimisation. However, their approach did not include semantic information to focus on target objects. Compared to these works, we propose a novel gradient-based approach for local NBV planning to perceive single target objects better, using volumetric and semantic information merged over multiple viewpoints.

In the agro-food domain, active vision methods have been shown to improve perception for tasks such as plant phenotyping \cite{gibbs2019active}, volume estimation of fruits \cite{zaenker2021viewpoint, marangoz2022fruit}, and mapping crop fields \cite{ruckin2022adaptive}. In our previous work, we showed that active vision can efficiently search and detect the task-relevant plant parts by focusing attention towards them \cite{burusa2022attention} and using semantic knowledge \cite{burusa2023efficient}. However, improving the perception of single target objects using active vision has not been widely studied.


\section{Problem formulation} \label{sec:formulation}

The common formulation of NBV planning methods is to find the next-best camera viewpoint to explore a bounded 3D space $V^T \subset \mathbb{R}^3$, to determine the free $V^T_\text{free} \subseteq V^T$ and occupied $V^T_\text{occ} \subseteq V^T$ regions, starting from $V^T$ being unknown. Since we are interested in perceiving only a target object, we define a region of interest (ROI) $V^T_\text{ROI} \subset V^T$, which is the region expected to contain the target object. We assume that the location of ROI within $V^T$ is given. Furthermore, we consider the semantic class labels and confidence scores from an object detection module to distinguish the target object from other objects. So, we modify the formulation of our NBV planning problem to finding the next-best camera viewpoint to explore $V^T_\text{ROI}$, to determine the free region $V^T_\text{free}$ and the region that belongs to the target object $V^T_\text{tar} \subset V^T_\text{ROI}$, starting from $V^T_\text{ROI}$ being unknown.

\subsection{Gradient-based optimisation}

We propose a gradient-based optimisation approach to the formulated problem. Our idea is to compute a viewpoint utility function in a differentiable way, so that its gradient can be computed with respect to the current viewpoint. Then, the viewpoint can be moved along the gradient direction to locally maximise this utility. The attributes of this gradient-ascent optimisation problem are defined as follows:

\subsubsection{Optimisation parameters} \label{sec:opti_params}
The optimisation parameters depended on the viewpoint. The viewpoint was defined using a camera position $p^c \in V^C \subset \mathbb{R}^3$, where $V^C \cap V^T = \emptyset$, and an expected target position $p^t \in V^T_\text{ROI}$. We constrained the viewpoint such that the camera was positioned at $p^c$ and oriented towards $p^t$. The roll of the camera was left uncontrolled, as changing it would not contribute much to the utility. So, $p^c$ and $p^t$ defined a 5 degrees-of-freedom (DoF) viewpoint and hence formed the optimisation parameters,
\begin{align}
    \xi = \{p^c, p^t\}.
\end{align}
The camera orientation $q^c$ can be extracted from $\xi$ as the rotation defined by the vector from $p^c$ to $p^t$ in the global frame. So, the complete camera viewpoint can be recovered as $\{p^c, q^c\}$.

\subsubsection{Objective function}
The objective function was the viewpoint utility that needed to be maximised. To perform gradient-ascent optimisation, the viewpoint utility $f(\xi)$ needs to be differentiable with respect to $\xi$. The details of the viewpoint utility are provided in Sec. \ref{sec:view_utility}.

\subsubsection{Optimisation problem}
The problem was to determine the viewpoint parameters that would maximise the expected viewpoint utility $f(\xi)$, that is,
\begin{align}
    \xi^* =& \;\underset{\xi}{\arg\max} \; f(\xi), \\
    \text{s.t.} \quad & p^c \in V^C, \; p^t \in V^T_\text{ROI},
\end{align}
where $V^C$ and $V^T_\text{ROI}$ defined the spatial constraints on the parameters. We hypothesise that such a gradient-based approach can smoothly guide the camera to perceive the target object better.


\section{Gradient-based local NBV planner}

\begin{figure*}[htbp]
     \centering
     \includegraphics[width=\textwidth]{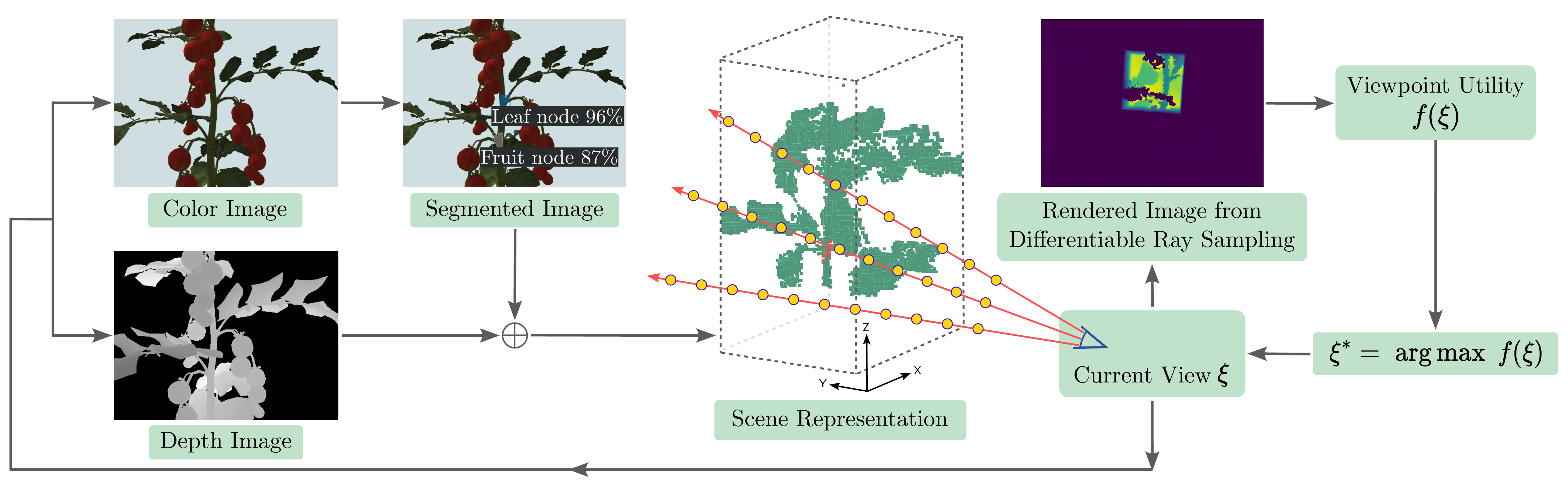}
     \caption{\small The pipeline of our proposed gradient-based NBV method. The nodes are detected in the color image using Mask R-CNN. The resulting segmented image, together with depth image, is inserted into the voxel grid to merge information from multiple views. The utility of the current view is computed using differentiable ray sampling, which provides a gradient along which the camera is moved.}
     \label{fig:gradient_nbv_pipeline}
\end{figure*}

To apply the method to our greenhouse scenario, we chose the target object as nodes of tomato plants, with the aim of reconstructing their 3D surface and estimating their 3D position. The major steps of our approach are as follows: (i) detect the node in the current viewpoint and estimate its position (Sec. \ref{sec:node_detection}), (ii) merge current information about the node with previously acquired information using a 4D scene representation (Sec. \ref{sec:voxel_grid}), (iii) compute the viewpoint utility of the current viewpoint and estimate the gradient (Sec. \ref{sec:view_utility}), and (iv) move in the direction of the gradient to potentially maximise the viewpoint utility (Sec. \ref{sec:view_planning}). We also keep track of the nodes over multiple viewpoints using a Kalman filter (Sec. \ref{sec:tracking}). The pipeline of the proposed method is illustrated in Fig. \ref{fig:gradient_nbv_pipeline}.

\subsection{Node detection and position estimation} \label{sec:node_detection}

The first step was to detect the node and extract information to estimate its 3D position. We achieved this by using an RGB-D camera, that provided color and depth images. A convolutional neural network, Mask R-CNN \cite{he2017mask} with a ResNet-50+FPN backbone, was used to perform instance segmentation on the color images and generate segmented images that separated the nodes from the background. We fine-tuned Mask R-CNN to detect nodes by training it on a custom dataset. The images had a dimension of $960\times540$ pixels and were collected in a simulated environment in Gazebo \cite{gazebosimGazebo}. 90 images were collected, of which 72 were used for training and 18 for validation. A low amount of training data was sufficient as there was not much variation between the nodes of the plants.

The segmented image, along with the depth image, were then used to estimate the 3D positions of the detected nodes. To do so, the depth information was converted to a point cloud that was aligned with the segmented image. For each detected node, its instance mask was applied to the aligned point cloud and the points that belonged to the detected node were extracted. The 3D position of the node was estimated as the mean position of the extracted points. The uncertainty in the 3D position was estimated as the variance of the extracted points along the viewing direction of the camera, as this was the main source of error in 3D position due to depth noise or error in the instance mask of Mask R-CNN.

\subsection{Scene representation with voxel grid} \label{sec:voxel_grid}

The depth and segmented images from the previous step represented the scene only from the current viewpoint. To represent the scene over multiple viewpoints, we used a 4D voxel grid $\mathcal{M} \in \mathbb{R}^{W \times H \times D \times C}$ of width $W$, height $H$, depth $D$, and channels $C$. The grid was within the bounded 3D space $V^T$. Each voxel in the grid had $C=4$ channels, containing the occupancy probability, semantic class label, semantic probability, and region-of-interest (ROI) indicator. These terms are explained below. The voxel grid is similar to an Octomap \cite{hornung2013octomap}, but extended to store multiple channels and made GPU-compatible for faster computation.

\subsubsection{Occupancy probability}
$p_o(x) \in [0,1]$ referred to the probability of a voxel $x$ being occupied. A probability of $p_o=0$ implied that the voxel was empty and $p_o=1$ implied that it is occupied. The values in between implied an uncertainty regarding the occupancy, with maximum uncertainty at $0.5$. All voxels were initialised with $p_o=0.5$ as there was no prior knowledge about their occupancy.

\subsubsection{Semantic class label and probability}
The semantic class label $c_s(x)$ referred to the type of node that a voxel $x$ belonged to, i.e. leaf or fruit, and the semantic probability $p_s(x) \in [0,1]$ referred to the probability of voxel $x$ belonging to the class $c_s(x)$. The voxels that did not belong to any node were considered as background. Hence, the class labels were defined as $c_s(x) \in \{-1=\text{background}, 0=\text{fruit node}, 1=\text{leaf node}\}$. A semantic probability of $p_s=0$ implied that the voxel did not belong to class $c_s(x)$, while $p_s=1$ implied that the voxel belonged to class $c_s(x)$. $c_s(x)$ was most uncertain when the semantic probability was $0.5$. All voxels, except the ones within the ROI, were initialised as background with a probability close to zero, since we wanted to focus only on the target object and ignore the rest of the plant.

\subsubsection{Region of interest (ROI) indicator}
It consisted of binary values to indicate if a voxel $x$ belonged to an ROI. The ROI was defined as an axis-aligned cube of edge length $0.06$m centered around a given position, which was a rough indication of where the target object was expected. It could either be obtained from a global NBV planner \cite{burusa2023efficient} or a predefined scan of the whole plant. The semantic probabilities of the voxels within the ROI were initialised as $p_s=0.5$ to guide the view planning to focus on the ROI.

With new depth and semantic measurements (Sec. \ref{sec:node_detection}), $p_o$ was updated using the probabilistic sensor fusion method proposed by \cite{hornung2013octomap}. The update of $c_s$ and $p_s$ had two cases. When the newly measured class label matched the previous estimate in a voxel, $p_s$ was updated similar to $p_o$. Else, the class label with the greatest semantic probability was accepted. The ROI did not change with new measurements.

\subsection{Viewpoint utility with differentiable ray sampling} \label{sec:view_utility}

The utility of a viewpoint $\xi=\{p^c, p^t\}$ was determined by estimating its semantic information gain $I_s(\xi)$, i.e., the amount of semantic information expected to be gained if the camera was moved there.

\subsubsection{Differentiable ray sampling} \label{sec:ray_sampling}
To compute $I_s(\xi)$, we first needed to extract the occupancy and semantic probabilities of the voxels within the field-of-view (FoV) of $\xi$. This was done using differentiable ray sampling. We cast a set of rays from the camera position $p^c$ along the camera FoV. A single ray $r$ was defined as,
\begin{align} \label{eq:ray}
    r(t) &= p^c + t \; d,
\end{align}
where $d$ was the viewing direction and $t$ limited the ray between the near $t_n=0.10$ and far $t_f=0.75$ bounds. Please note that the viewing direction $d$ depends on the camera orientation $q^c$ and hence $r$ is differentiable with respect to $\xi$. A set of $N_r=128$ points were uniformly sampled along each ray. For each sampled point, the occupancy and semantic probabilities were read from the voxel grid $\mathcal{M}$ by identifying the closest voxels to the sampled point and performing a trilinear interpolation of the probabilities of those voxels. The sampled probabilities were differentiable with respect to $\xi$, as the interpolation operation was implemented in a differentiable way, similar to \cite{jaderberg2015spatial}.

\subsubsection{Expected semantic information gain}
Using the sampled occupancy and semantic probabilities along a ray $r$, we compute the expected semantic information gain $I_s(r)$ for the ray as, 
\begin{align}
    I_s(r) &= \sum_{i=1}^{N_r} T(i)I_s(i), \\
    \text{where} \quad T(i) &= \prod_{j=1}^{i-1} (1 - p_o(j)).
\end{align}
Here, the function $I_s(i)$ denotes the semantic information expected to be gained by observing the sampled point $i$ along the ray $r$. It was defined by Shannon's entropy,
\begin{align}
    I_s(i) &= -p_s(i)\log_2(p_s(i)) - (1 - p_s(i))\log_2(1 - p_s(i)).
\end{align}
The function $T(i)$ is the accumulated transmittance along the ray until point $i$, i.e., the probability that the ray passed without hitting an occlusion. Intuitively, $I_s(r)$ is the sum of the expected semantic information gains of points along the ray $r$, which only considers the points that are expected to be visible from viewpoint $\xi$. The rest of the points, that are expected to be occluded, will have a low transmittance value and hence will be omitted. The total expected semantic information gain $I_s$ for $\xi$ was then obtained by summing the expected gains along all rays $\mathcal{R}$,
\begin{align}
    I_s(\xi) &= \sum_{r \in \mathcal{R}} I_s(r).
\end{align}

\subsubsection{Gradient of the viewpoint utility}
The gradient was obtained by differentiating the expected semantic information gain $I_s(\xi)$ with respect to the viewpoint $\xi$. The gradient computation was handled by the auto-differentiation feature of the PyTorch \cite{paszke2019pytorch} library.

\subsection{Viewpoint planning with gradient ascent} \label{sec:view_planning}

Once the gradient was computed, the viewpoint was moved in the direction of the gradient so that $I_s(\xi)$ could be maximised locally. This gradient-ascent step was scaled by a step size $\alpha$, which was a hyperparameter.
\begin{align}
    \xi_{k+1} &= \xi_k + \alpha \frac{\partial I_s(\xi_k)}{\partial \xi_k}
\end{align}

\subsection{Tracking nodes over multiple viewpoints} \label{sec:tracking}

Since multiple nodes are detected by Mask R-CNN across multiple views, we needed a tracking method to properly associate the detected nodes in the current view with nodes from previous views, so that the position estimate of the target node could be correctly updated. We achieved this using a Kalman filter with the 3D positions and class labels of nodes as states. Assuming that there were $N$ distinct nodes that were previously detected, the Kalman filter state at step $k$ was defined as $S_k = \{o_k^1, o_k^2,...,o_k^N\}$, where $o_k^j = \{p_k^j, c_k^j\}$ denoted the node $j$ and consisted of its 3D position $p_k^j$ and semantic class label $c_k^j$. The 3D position was defined with a Gaussian mean and covariance matrix, $p_k^j = \{\mu_k^j, \Sigma_k^j\}$. For targeted perception, one node from $S_k$ was selected as the target object and its position was assigned to $p^t$.

\subsubsection{State prediction}
In the prediction step, both the position and class labels of the nodes were kept constant, as we assumed the scene to be static.

\subsubsection{State update} 
The 3D position was updated through a regular Kalman filter update step which tries to minimise the mean squared error between the measured and estimated positions. For the class labels, we used a majority-voting-based approach, where the label that was observed in most of the viewpoints was assigned to the node.


\section{Methods for comparison}

We compared our gradient-based optimization approach with a sampling-based approach for local viewpoint planning. For the sampling-based methods, a set of $N_c$ candidate viewpoints were locally sampled close to the current viewpoint. In particular, we uniformly sampled two sets of $N_c$ points within a radius of $0.1$m from $p^c$ and $p^t$ of the current viewpoint $\xi$. These pairs of sampled points were used to define $N_c$ 5-DoF candidates, using the procedure discussed in Sec. \ref{sec:opti_params}. The next camera viewpoint was selected from the candidates using two different approaches:

\subsubsection{Sampling-based semantic NBV planner}
The semantic NBV planner estimated the semantic information gain $I_s(\xi)$ for all candidates and picked the one that maximised the gain. $I_s(\xi)$ was estimated according to the ray sampling procedure discussed in Sec. \ref{sec:ray_sampling}. The semantic NBV planner is an effective algorithm for planning viewpoints to improve the semantic information in a scene or a set of objects \cite{burusa2023efficient}.

\subsubsection{Sampling-based random planner}
The random planner picked a viewpoint at random from the set of candidates. This planner was used as a baseline to verify that our method worked better than random selection.

In the following sections, we refer to these planners as `SamplingNBV' and `Random' respectively, and we refer to our proposed planner as `GradientNBV'.


\section{Experiments and Results}

\subsection{Simulation setup}
The robot consisted of a 6-DoF manipulator (ABB IRB 1200) with an RGB-D camera (Intel RealSense L515) attached to it. The images obtained from the RGB-D camera had a resolution of $960\times540$ pixels. The resolution of the voxel grid was $0.002$m and its dimensions were $0.3\times0.3\times0.7$m$^3$. This dimension can be varied and does not constraint the algorithm. The step size for GradientNBV was set to $0.065$, which roughly moved the camera the same distance as SamplingNBV for the initial steps. We used eight 3D mesh models of tomato plants of varying growth stages and structural complexity.

\subsection{Evaluation metrics}

We used multiple metrics to evaluate the planners on the accuracy of 3D reconstruction, node detection, position estimation, and the efficiency of view planning.

\subsubsection{ROI coverage}
It measured how completely an ROI was observed by a planner. It was defined as the percentage of voxels within the ROI that were viewed by the camera, at least from one viewpoint, out of all voxels in the ROI.

\subsubsection{F1-score of 3D node reconstruction}
It measured how complete and accurate the node reconstruction was. It compared the reconstructed point cloud from view planning with the ground-truth point cloud from the true 3D mesh of the plants. Only the region within a 6cm cube around the true position of the node was evaluated, which was the volume covered by the target ROI. A reconstructed point was considered true positive when it was within $0.002$m from the ground-truth point, based on the resolution of the voxel grid.

\subsubsection{Number of ray-tracing calls}
It indicated the computational cost of the view planners since the ray-tracing operation was the most computationally expensive step. The time for one ray-tracing call was $0.06$ seconds.

\subsubsection{Trajectory distance}
The efficiency of the generated sequence of viewpoints was estimated by summing the Euclidean distances between two consecutive viewpoints along the trajectories.

\subsubsection{Recall of occluded node detection}
It measured the percentage of nodes that were detected accurately among the subset of nodes that were occluded and undetected at the starting view. A node was considered as accurately detected when its position was within $0.02$m from the ground-truth and its detected class label matched the true class label.

\subsubsection{Standard deviation $(\sigma$) of 3D node position}
It was the estimated uncertainty on the node position from the Kalman filter. A low standard deviation indicated more confidence in the position estimation.

\subsection{Results of occlusion-handling behaviour}

In this experiment, we analysed if the trajectory generated by the viewpoint planners handled occlusions in a desirable way. We placed a single target object in front of the robot, with a variation of $\pm0.15$m in the y and z axes. An ROI of size $0.06$m was defined around it. We then partially occluded the target object from the camera's view using a box and tested if the view planners could plan an efficient set of viewpoints that maximised the ROI coverage. Four cases were considered, where the left, right, top, or bottom part of the target object was occluded. For each case, four different initial values were randomly assigned to $\xi=\{p^c, p^t\}$, with a total of 16 trials per planner. The planning was terminated at the end of $20$ viewpoints.

\definecolor{Green}{rgb}{0.753,0.886,0.792}
\begin{table*}[htbp]
\centering
\caption{\small Results for the occlusion-handling behaviour. The average performance across 16 experiments is shown. We observed that GradientNBV and SamplingNBV were able to explore the target ROI equally well, but the GradientNBV used ten times less ray-tracing calls and generated four times more efficient trajectories. View 0 was predefined.}
\label{tab:occlusion_handling}
\renewcommand{\arraystretch}{1.1}
\setlength\tabcolsep{7pt}
\begin{tabular}{l | c c c c c | c c c c c | c c c c c}
\hline
\rowcolor{Green} & \multicolumn{5}{c|}{ROI coverage (\%) $\uparrow$} & \multicolumn{5}{c|}{Number of ray-tracing calls (\#) $\downarrow$} & \multicolumn{5}{c}{Trajectory distance (m) $\downarrow$} \\
\hline
\rowcolor{Green} \# Viewpoints & 0 & 5 & 10 & 15 & 20 & 0 & 5 & 10 & 15 & 20 & 0 & 5 & 10 & 15 & 20 \\
\hline
GradientNBV & 10.2 & \textbf{75.2} & 89.7 & 92.6 & 93.8 & 0 & \textbf{5} & \textbf{10} & \textbf{15} & \textbf{20} & 0.00 & \textbf{0.15} & \textbf{0.23} & \textbf{0.28} & \textbf{0.31} \\
SamplingNBV & 12.9 & 72.3 & \textbf{90.3} & \textbf{93.7} & \textbf{95.1} & 0 & 50 & 100 & 150 & 200 & 0.00 & 0.45 & 0.84 & 1.25 & 1.67 \\
Random & 7.3 & 45.1 & 58.0 & 63.7 & 70.4 & - & - & - & - & - & 0.00 & 0.40 & 0.85 & 1.29 & 1.70 \\
\hline
\end{tabular}
\end{table*}

\begin{figure}[htbp]
     \centering
     \includegraphics[width=0.48\textwidth]{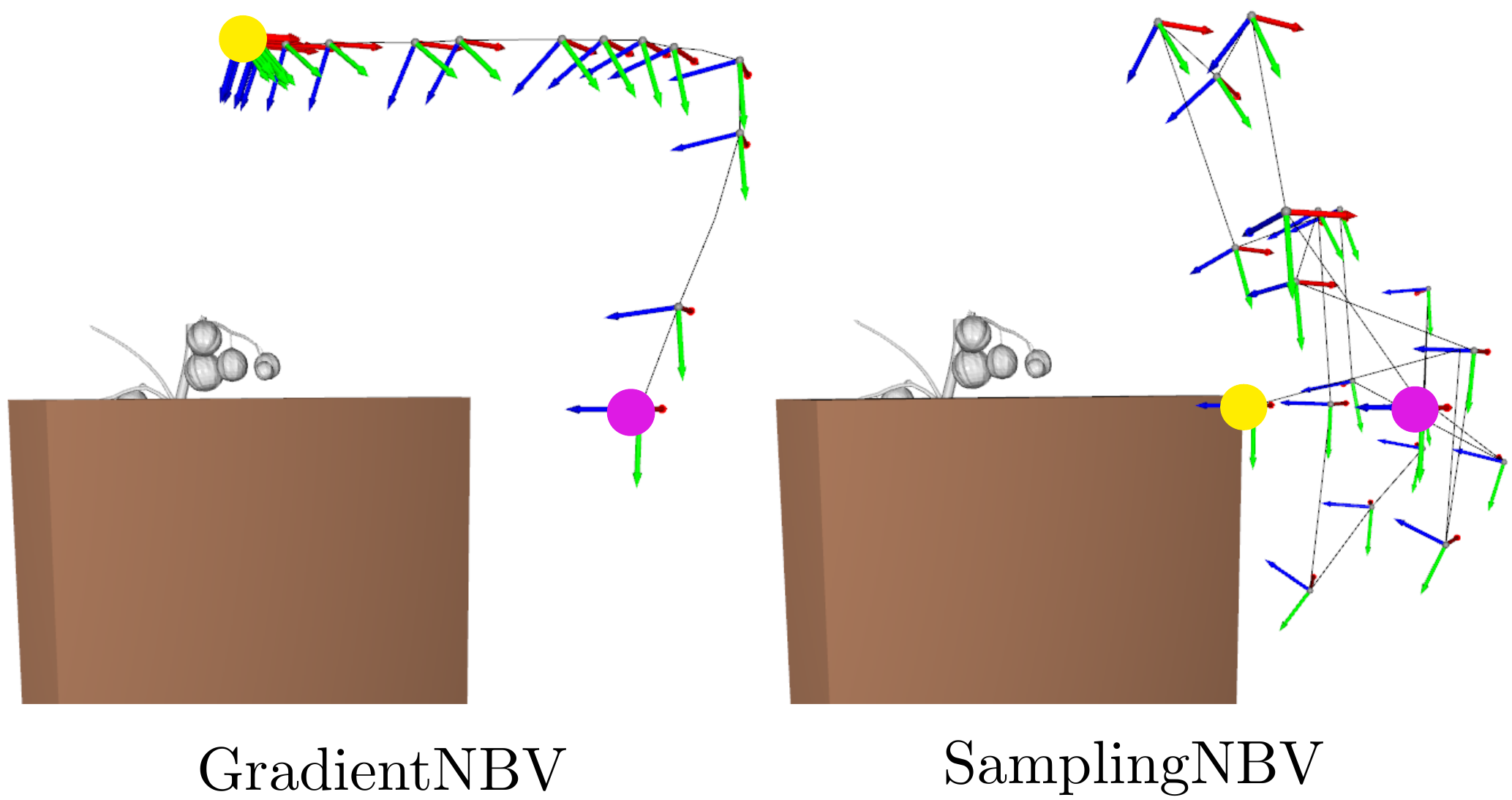}
     \caption{\small Qualitative analysis of the trajectories generated by the viewpoint planners. The blue axis shows the viewing direction of the camera. The start and end viewpoints are marked in pink and yellow respectively.}
     \label{fig:trajectory_analysis}
\end{figure}

\begin{table*}[htbp]
\centering
\caption{\small Results for node reconstruction and position estimation of plant nodes within a target ROI. We observed that the GradientNBV and SamplingNBV explored the target ROI equally well, but the GradientNBV used ten times less ray-tracing calls and generated $28\%$ more efficient trajectories. The average performance across $288$ experiments is shown. View 0 was predefined.}
\label{tab:node_perception}
\renewcommand{\arraystretch}{1.1}
\setlength\tabcolsep{6pt}
\begin{tabular}{l | c c c c c c | c c c c c c | c c c c c c}
\hline
\rowcolor{Green} & \multicolumn{6}{c|}{ROI coverage (\%) $\uparrow$} & \multicolumn{6}{c|}{F1-score of 3D node reconstruction (\%) $\uparrow$} & \multicolumn{6}{c}{Number of ray-tracing calls (\#) $\downarrow$} \\
\hline
\rowcolor{Green} \# Viewpoints & 0 & 1 & 2 & 3 & 4 & 5 & 0 & 1 & 2 & 3 & 4 & 5 & 0 & 1 & 2 & 3 & 4 & 5 \\
\hline
GradientNBV & 45.0 & \textbf{65.2} & \textbf{74.4} & \textbf{79.6} & 82.9 & 84.8 & 72.1 & \textbf{82.0} & \textbf{84.6} & 86.0 & 86.8 & 87.3 & 0 & \textbf{1} & \textbf{2} & \textbf{3} & \textbf{4} & \textbf{5} \\
SamplingNBV & 45.1 & 64.1 & 73.7 & 79.5 & \textbf{83.2} & \textbf{86.1} & 71.9 & 80.5 & 84.1 & \textbf{86.1} & \textbf{87.4} & \textbf{88.5} & 0 & 10 & 20 & 30 & 40 & 50 \\
Random & 44.7 & 58.0 & 65.9 & 70.3 & 73.8 & 76.0 & 71.9 & 77.9 & 80.6 & 82.1 & 83.0 & 83.7 & - & - & - & - & - & - \\
\hline
\hline
\rowcolor{Green} & \multicolumn{6}{c|}{Trajectory distance (m) $\downarrow$} & \multicolumn{6}{c|}{Recall of occluded node detection (\%) $\uparrow$} & \multicolumn{6}{c}{$\sigma$ of 3D node position (m$\times10^{-2}$) $\downarrow$} \\
\hline
\rowcolor{Green} \# Viewpoints & 0 & 1 & 2 & 3 & 4 & 5 & 0 & 1 & 2 & 3 & 4 & 5 & 0 & 1 & 2 & 3 & 4 & 5 \\
\hline
GradientNBV & 0.0 & 0.11 & \textbf{0.18} & \textbf{0.23} & \textbf{0.27} & \textbf{0.31} & 0.0 & \textbf{18.2} & \textbf{28.5} & \textbf{35.2} & \textbf{39.4} & \textbf{41.8} & 1.7 & 1.6 & \textbf{1.4} & \textbf{1.3} & \textbf{1.3} & \textbf{1.2} \\
SamplingNBV & 0.0 & \textbf{0.10} & 0.19 & 0.28 & 0.35 & 0.43 & 0.0 & 7.5 & 16.2 & 20.6 & 24.4 & 29.4 & 1.7 & 1.6 & 1.5 & 1.4 & 1.4 & 1.3 \\
Random & 0.0 & \textbf{0.10} & 0.19 & 0.28 & 0.36 & 0.44 & 0.0 & 8.4 & 14.4 & 16.3 & 18.7 & 20.5 & 1.7 & 1.6 & 1.5 & 1.5 & 1.5 & 1.4 \\
\hline
\end{tabular}
\end{table*}

We compared the performance of the different planners in Table \ref{tab:occlusion_handling}. There was no significant difference in the performance of GradientNBV and SamplingNBV in terms of ROI coverage and F1-score of 3D node reconstruction. Also, both planners could effectively search and find the target object despite uncertainty in the target location. However, the SamplingNBV required ten times more ray-tracing calls to reach the same performance as it had to evaluate all candidate viewpoints before determining the next-best viewpoint. Moreover, the viewpoint trajectories generated by GradientNBV were three times more efficient compared to SamplingNBV and Random planners in terms of the average trajectory distance. The trajectories were also qualitatively analysed, as shown in Fig. \ref{fig:trajectory_analysis}. We found that the trajectories generated by GradientNBV were smoother and more efficient compared to SamplingNBV and Random planners.


\subsection{Results of node reconstruction and position estimation}

\setlength{\belowcaptionskip}{-10pt}
\begin{figure}[htbp]
     \centering
     \includegraphics[width=0.48\textwidth]{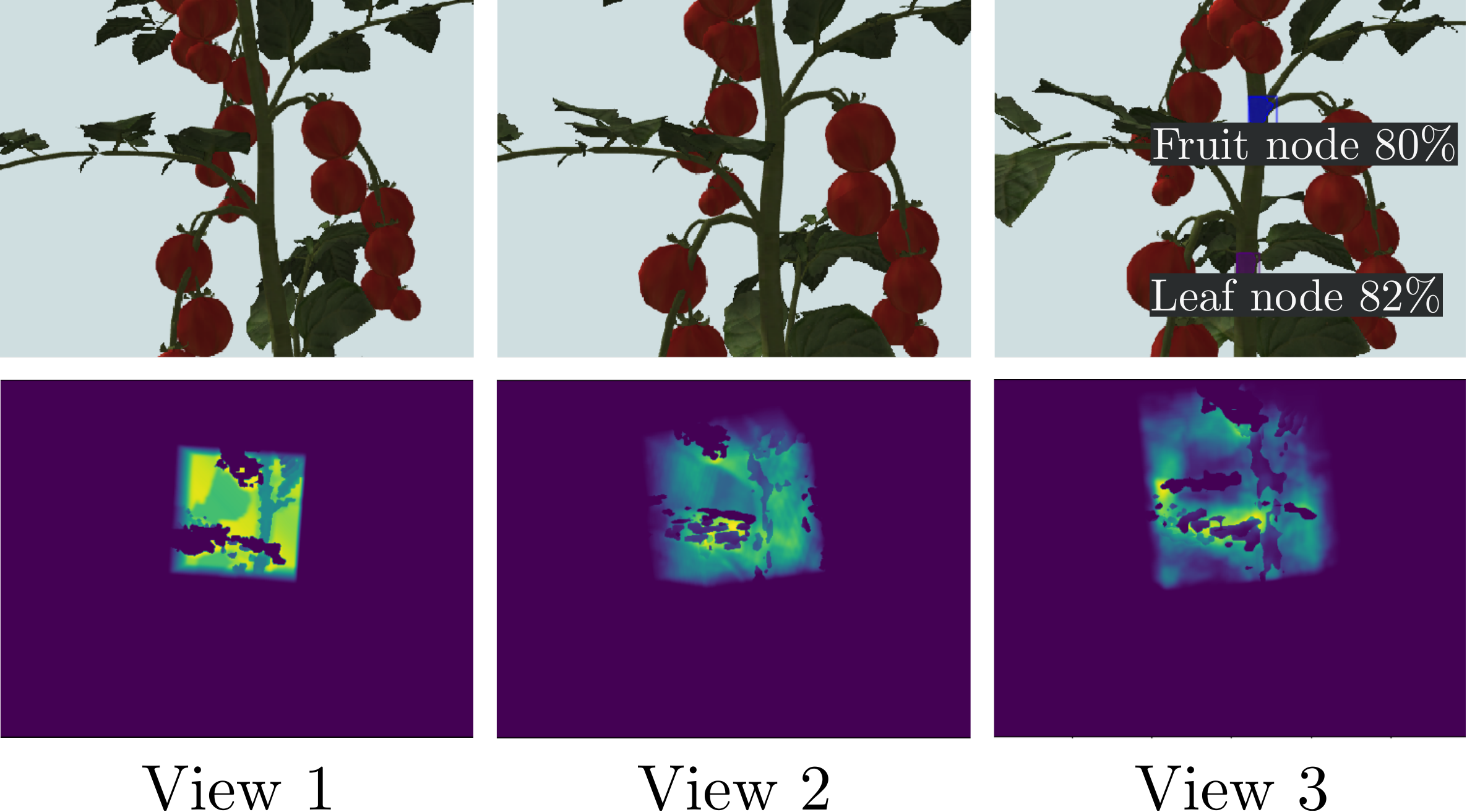}
     \caption{\small Example of three consecutive views from GradientNBV. The top row shows the segmented images and the bottom row shows the viewpoint utility rendered from the current view (yellow: high, blue: low). We can observe that the nodes were detected in View 3 as the camera moved closer. The utility of the viewpoints within the ROI reduced with each viewpoint.}
     \label{fig:gain_visualisation}
\end{figure}

In this experiment, we tested how well the gradient-based NBV planner handled occlusions in plants. In particular, we analysed if the planner could improve the 3D reconstruction and position estimation of a target node on the plant. We placed a plant in front of the robot with an uncertainty of $\pm10$cm in the y and z axes. An arbitrary node was selected either at the bottom, middle, or top of the plant. An ROI of size $6$cm was defined around this target node with an uncertainty of $\pm3$cm, so that the true position of the node was still within the ROI. $p^t$ was initialised with the target node's position and $p^c$ was predefined close to it. This setup mimicked the situation in a greenhouse where the 3D position of a node was roughly known, either through direct observation or expectation, and the robot aimed to improve its accuracy for harvesting or de-leafing. Hence, the objective was to plan a set of viewpoints that would improve the perception of the target node. Multiple experiments were conducted using $8$ plants with $3$ target nodes and $12$ rotations along the z-axis per plant, leading to a total of $288$ trials per planner. In $165$ trials, the nodes were occluded and undetected in the starting view. The planning was terminated at the end of $5$ viewpoints.

In Table \ref{tab:node_perception}, we quantitatively analysed the planner performances. We observed that the GradientNBV and SamplingNBV planners performed equally well in ROI coverage and F1-score, which indicated that both planners were able to handle occlusion and improve the 3D reconstruction of the target nodes. However, GradientNBV was ten times more computationally efficient and $28\%$ more efficient in terms of the camera trajectory. Regarding the detection of occluded nodes, the recall of GradientNBV was $13\%$ more than SamplingNBV and $22\%$ more than Random, indicating that the GradientNBV was more effective in detecting the nodes that were undetected in the starting view. The estimated standard deviation of the 3D node positions were low and reduced with more viewpoints for all the planners. Fig. \ref{fig:gain_visualisation} visualises the node detection and the rendered utility for views generated by GradientNBV.


\section{Conclusion and Future work}

In this paper, we presented a novel gradient-based NBV planning method using differentiable ray sampling for improving the perception of target objects in occluded scenarios. Compared to previous works on NBV planning, our approach removes the need to sample and evaluate multiple candidate viewpoints. We performed simulation experiments with 3D mesh models of tomato plants to evaluate the performance of our planner on the tasks of 3D reconstruction and position estimation of nodes. Our gradient-based NBV planner was able to explore the target region, reconstruct the nodes, and estimate their 3D position as accurately as a sampling-based NBV planner, while taking ten times less computation and generating $28\%$ more efficient trajectories. Our results clearly show the advantage of using the gradient-based NBV planner for local viewpoint planning in occluded scenarios. Our approach can help robots to efficiently perceive a target plant node for grasping or cutting, and contribute to improved harvesting and de-leafing in greenhouses. In future work, we can extend the method for global planning in larger scenarios.


\newpage
\bibliographystyle{IEEEtran}
\bibliography{IEEErefs}


\end{document}